\documentclass[
]{ceurart}

\graphicspath{ {./images/} }
\usepackage{array}
\usepackage{hyperref}

\begin{document}

\copyrightyear{2021}
\copyrightclause{Copyright for this paper by its authors.
  Use permitted under Creative Commons License Attribution 4.0
  International (CC BY 4.0).}

\conference{Related - The 1st International Workshop of Relations in the Legal Domain, ICAIL 2021 – The 18th International Conference on Artificial Intelligence and Law, University of São Paulo, Monday June 21 – Friday June 25, 2021}

\title{Classification of Contract-Amendment Relationships}

\author[1]{Fuqi Song}[email=fsong@hyperlex.ai]
\address[1]{Data Science, Hyperlex, 13 Rue de la Grange Batelière, 75009 Paris, France}

\begin{abstract}
  In Contract Life-cycle Management (\textit{CLM}), managing and tracking the master agreements and their associated amendments is essential, in order to be kept informed with different due dates and obligations. An automatic solution can facilitate the daily jobs and improve the efficiency of legal practitioners. In this paper, we propose an approach based on machine learning (ML) and Natural Language Processing (NLP) to detect the amendment relationship between two documents. The algorithm takes two PDF documents preprocessed by OCR (\textit{Optical Character Recognition}) and NER (\textit{Named Entity Recognition}) as input, and then it builds the features of each document pair and classifies the relationship. We experimented with different configurations on a dataset consisting of 1124 pairs of contract-amendment documents in English and French. The best result obtained a F1-score of 91\%, which outperformed 23\% compared to a heuristic-based baseline.
\end{abstract}


\maketitle

\section{Introduction and Problem Statement}
\label{sec:introduction}
In Contract Life-cycle Management(\textit{CLM}), the contracts and other documents are not isolated elements. There exists links among them, the most common and important one being the contract-amendment relationship between a master agreement (\textit{MA}) and an amendment. Tracking and handling such links is essential in different CLM tasks so as to lower the potential legal risks and be up to date relatively to the evolution of a contract through its amendments. 
Conventionally, the task is performed manually or semi-automatically within a digital solution, which is time-consuming and error-prone.
A fully automatic solution is expected to overcome these drawbacks and to facilitate the CLM process.

In this article, we therefore propose a method for automatically detecting linked documents based on machine learning algorithms and NLP techniques. 
The problem can be formulated as a binary classification problem that takes two documents as input and classifies the relationship between them. A key problem is to identify a good feature set for the classification algorithms. 
We apply a ML-driven preprocessing pipeline, including principally OCR and NER. The pipeline outputs the recognized document content and named entities, such as corporate names and contract numbers. 
Depending on the quality and content of documents, the extracted text and named entities might contain errors and be inaccurate. 
Taking these factors into account and trying to be robust, we propose a similarity and cross reference-based approach for extracting features from a pair of documents. 
The approach is robust to different errors introduced during preprocessing and allows taking into account multiple uncertain factors to classify the relationship. 

\begin{figure}
  \centering
  \includegraphics[width=12cm]{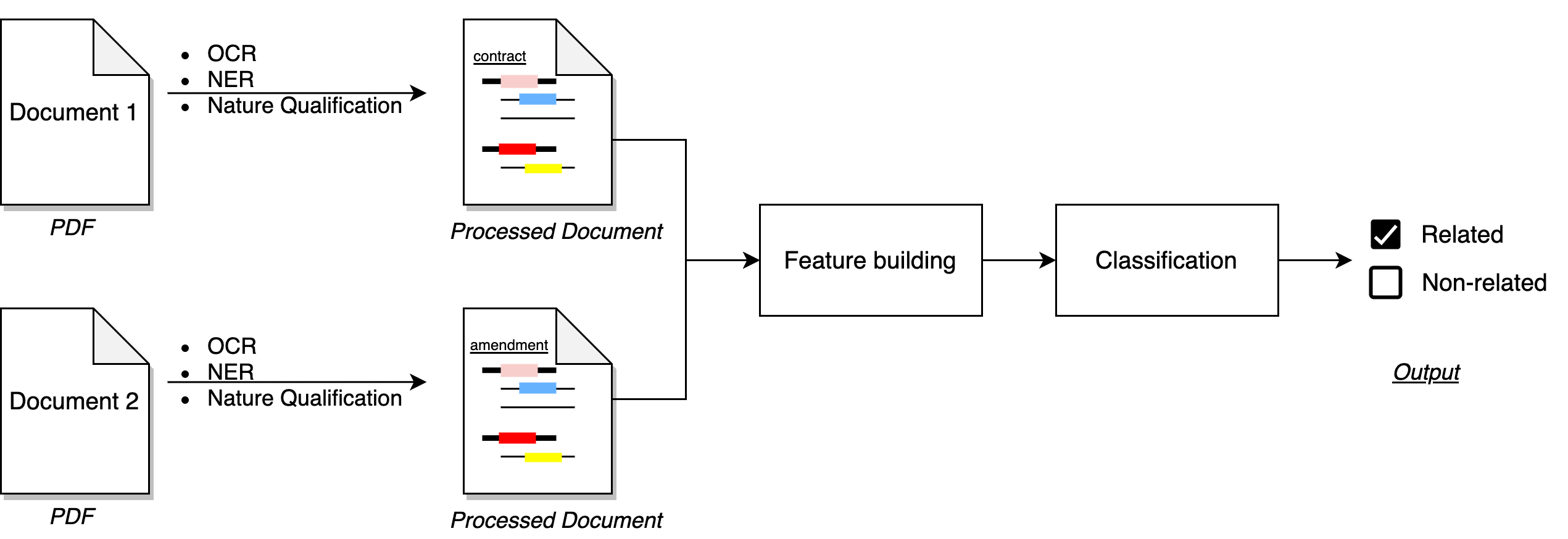}
  \caption{General pipeline of contract-amendment relationship classification}
  \label{fig1:general_schema}
\end{figure}

The general schema of the entire process is illustrated in Figure~\ref{fig1:general_schema}: we take a pair of PDF documents as input and perform three main steps to detect whether or not the two documents are related, namely, preprocessing, feature extraction, and classification. 
The paper focuses on the definition of our features and on the classification algorithms. 
The preprocessing step that applies OCR (\textit{Optical Character Recognition}), NER (\textit{Named Entity Recognition}~\cite{Nadeau2007,Yadav2019}) and entity aggregation~\cite{Pons2006,Fortunato2010} will not be elaborated. 

The rest of the paper is organized as follows: Section~\ref{sec:feature_building} analyzes the key features that can distinguish the related contract-amendment documents from nonrelated ones and explains how the features are represented. 
Section~\ref{sec:baseline} presents the dataset and the baseline algorithm used to evaluate our approach. 
Section~\ref{sec:classification} experiments different configurations to classify the relationships and analyzes the benchmarking results.
Section~\ref{sec:applicaiton} discusses two typical application scenarios using the contract-amendment classification as a key component.
Section~\ref{sec:conclusion} draws some conclusions and extends the future works.

\section{Feature Building}
\label{sec:feature_building}

\subsection{Analysis}
\label{sec:feature_analysis}

In the quantification of the contract-amendment relationship of two documents, we have identified the following key pieces of information allowing to distinguish the related documents from nonrelated ones:

\begin{itemize}
    \item {\verb|Document name|}: In general, the document name provides a lot of indices that help to deduce the relationship between a pair of documents. Indeed, often the document name follows certain patterns (which vary for different persons and different organizations), for instance, \textit{Contract No. X12345.pdf} and \textit{Contract No. X12345 Amendment 1.pdf};

    \item {\verb|Legal parties|}: Very frequently, the legal parties engaged in the contract are the same for the master agreement and its amendments, with the roles of legal parties in a contract explained in~\cite{Song2020};

    \item {\verb|Document body|}: The document body of the master contract and amendments tend to be similar in general and are semantically related. An amendment recalls the key information in the master contract and specifies the modifications in relation to the master contract;

    \item {\verb|References|}: The indices that are referred explicitly in two documents to establish the relationship. The typical ones are dates and contract numbers, for instance, ``\textit{... Contract \textbf{N°X12345} signed on \textbf{14 May 2003} ...}" is a typical way used in an amendment to address the relationship with the master agreement.
\end{itemize}

Once we have identified the features, the next question is how to represent these features with numerical values.
One of the key issues is that the extracted information is not 100\% accurate, for instance, the extracted dates or legal parties might be inaccurate or missing. 
Therefore, we propose to build the features based on the distance between two pieces of information in two documents. 
Section~\ref{sec:document_representation} explains the representation of a single document and Section~\ref{sec:feature_representation} illustrates the feature representation of a document pair that will be used to classify the relationship.

\subsection{Document Representation}
\label{sec:document_representation}

A preprocessed document is formally denoted by:
\[doc = (\mbox{name}, \mbox{text}, \mbox{legal\_parties}, \mbox{keywords}, \mbox{nature}) \]
wherein 
\begin{itemize}
    \item \textit{name} denotes the file name given by users; 

    \item \textit{text} denotes the plain text extracted by OCR;

    \item \textit{legal\_parties} lists all distinct corporate names extracted by NER from the clause of declaration of parties; 

    \item \textit{keywords} includes named entities extracted by NER that could be used as cross references, and more specifically \textit{dates} and \textit{contract identifiers} in this paper.
    It is however important to note that the same entity may play distinct roles in the master agreement and in the amendment.
    For instance, a named entity \textit{signature date} in master agreement that is used as a reference in amendment can be with simple type \textit{date};

    \item \textit{nature} represents the type of a document in three categories: \textit{contract}, \textit{amendment} or \textit{other}. 
    \textit{nature} is determined during the preprocessing, thanks to a text classification algorithm (with F1-score about 90\%). 
    This information is used to filter the document pairs to classify. 
    More precisely, we classify the relationships only between a pair of documents where one is a \textit{contract} and the other is an \textit{amendment}.

\end{itemize}

\subsection{Feature Representation}
\label{sec:feature_representation}

The feature associated with a pair of documents $(doc_1, doc_2)$ is denoted as $ \mathcal{F} = (f_1, f_2, f_3, f_4) $ wherein:

\begin{itemize}
    \item $f_1$ {\verb|(Document name)|}: $f_1$ represents the similarity between the document names, the string metric is a string and token-based compound metric described in \cite{Song2020};
    
    \item $f_2$ {\verb|(Document text)|}: To compute the similarity of two texts, the first step is to embed the text to numerical vectors and then compute the cosine similarity \cite{schutze2008introduction}. 
    In this article, we test two embedding methods: TF-IDF and FastText, which is elaborated in Section~\ref{sec:classification};
    
    \item $f_3$ {\verb|(Legal parties)|}: The absolute number of shared legal parties. 
    For each corporate name in $doc_1.legal\_parties $, we compute the string similarity with each corporate name in $doc_2.legal\_parties $. 
    When the similarity is greater than the defined threshold (0.85), we increment the number of shared legal parties;
    
    \item $f_4$ {\verb|(References)|}: The absolute number of shared keywords, computed following the same principle as $f_3$.
\end{itemize}

Features $f_1$ and $f_2$ are real numbers ranging $[0,1]$ whereas features $f_3$ and $f4$ are discrete numbers $(0, 1, 2, ...)$.

\section{Dataset and Baseline}
\label{sec:baseline}
\label{sec:dataset}

The dataset\footnote{Due to confidentiality reasons, the dataset is not publicly accessible.} consists of 1124 pairs of documents in relationship \textit{contract-amendment} of real contracts from different companies. 
The dataset has been annotated manually by legal experts by presenting them the pairs of potentially linked documents.
The dataset includes different types of contracts with different levels of qualities. 
617 pairs of documents are in French and 507 pairs in English in order to test a robust multilingual approach.
As for the negative samples, 1124 document pairs have been sampled randomly from the contract base and verified manually by legal experts. 
The measurements are \textit{precision}, \textit{recall} and \textit{F1-score (macro)} \cite{powers2020evaluation}.

\begin{figure}
  \centering
  \includegraphics[width=12cm]{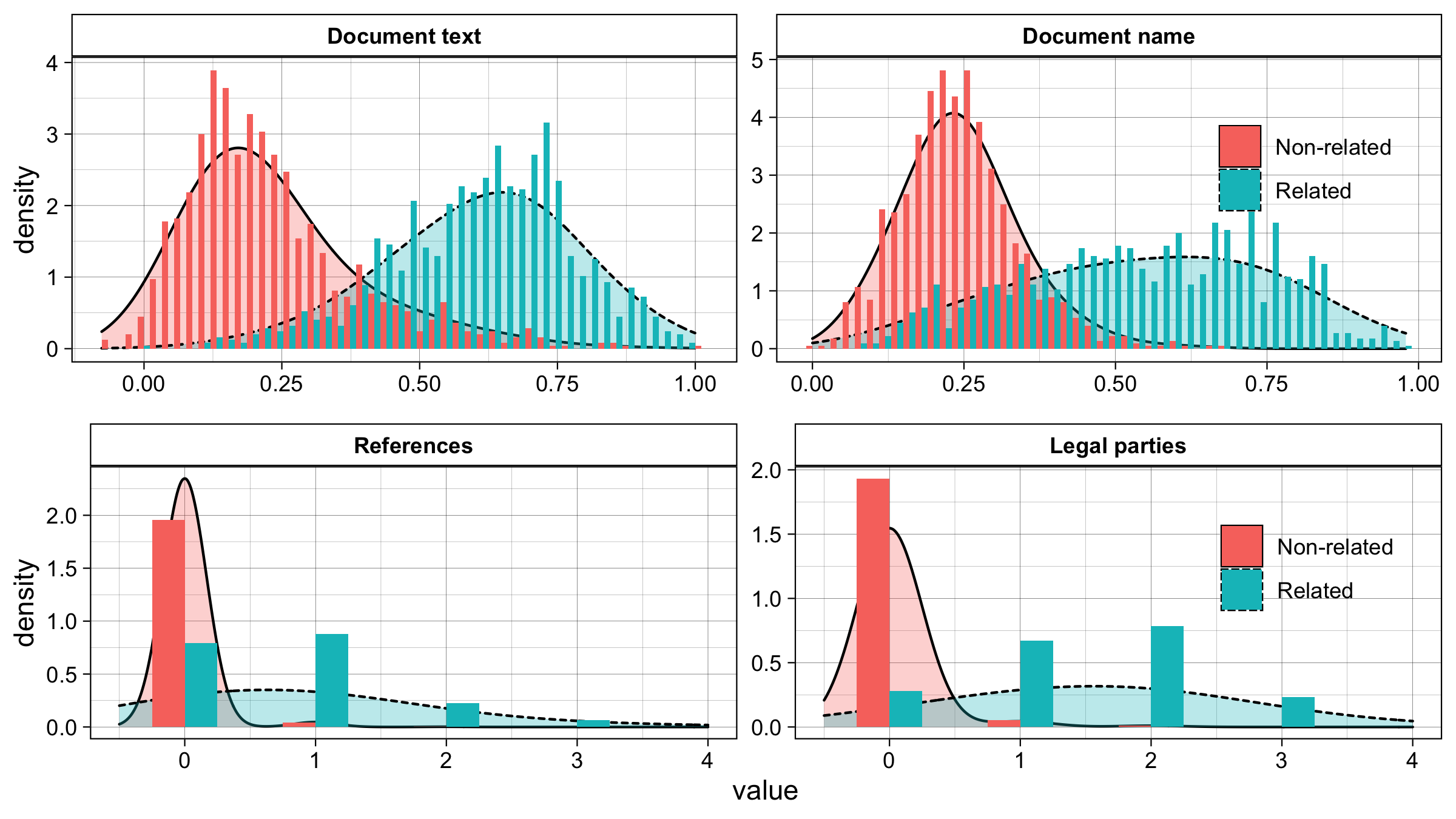}
  \caption{Distribution of feature values in relation to link types}
  \label{fig2:feature_distribution}
\end{figure}

The dataset is preprocessed using the pipeline illustrated in Figure~\ref{fig1:general_schema} which outputs the processed documents in the format defined in Section~\ref{sec:feature_representation}. 
The correlation between the selected features and the relationships to classify is analyzed in Figure~\ref{fig2:feature_distribution}. 
The figure shows the histogram and density of each feature in relation to the link types. We use TF-IDF as embedding for computing the text similarity. 
We observe a clear pattern between the related and nonrelated datasets on the four features, more precisely the values being generally higher for related pairs. However no feature is discriminating enough to separate related pairs from nonrelated ones.

To our knowledge, due to the specificity of the research problem, few works have been published on the topic of classification of contract-amendment relationships. 
To evaluate the proposed approach, we propose a heuristic-based baseline without the application of ML techniques, namely, using only the document name and the extracted text. 
The rules are as follows: if the similarities of document name and text (with TF-IDF) between two documents are both greater than 0.5 (as observed in Figure~\ref{fig2:feature_distribution}), we consider that the two documents are related, otherwise nonrelated.

\section{Classification}
\label{sec:classification}

We experimented with different configurations to evaluate the impacts of these variables on the classification and to try to find the best configuration for the final model.

\begin{itemize}
    \item {\verb|Text embedding|}: \textbf{TF-IDF} \cite{rajaraman_ullman_2011} and \textbf{FastText} \cite{bojanowski2017enriching} for evaluating the impacts on the document content feature. 
    FastText is a non-contextual word embedding taking the subwords into account, which is potentially more robust to OCR errors. 
    We use the pretrained English and French word vectors~\footnote{\url{https://fasttext.cc/docs/en/crawl-vectors.html}} for our experiments;

    \item {\verb|Transformations|}: The strategies to transform the feature values: 1) \textbf{None}: no transformation, 2) \textbf{Binary}: the real values are mapped to binary 0 or 1 relatively to a threshold, and 3) \textbf{Decimal}: the real values are mapped proportionally to an integer between 0 and 10;

    \item {\verb|Classification algorithms|}: 1) Random Forest (\textbf{RF}), 2) Linear SGD classifier (\textbf{Linear}), and 3) Multi-layer perceptron (\textbf{MLP}).

\end{itemize}

\begin{table}[ht]
\begin{center}
\caption{Benchmarking results of different configurations on test set}
\label{table:results}

\begin{tabular}{ lll
>{\em}c
>{\em}c
>{\em}c } 
\hline

\textbf{Classifier    } & 
\textbf{Embedding    } & 
\textbf{Transformation  } & 
\textbf{  Precision (\%) } & 
\textbf{  Recall (\%) } & 
\textbf{  F1-score (\%) } \\
\hline

\textbf{Baseline} & \textbf{TF-IDF} & \textbf{None} & \textbf{77.5} & \textbf{64.7} & \textbf{67.6} \\
\hline

RF & FastText & Decimal & 90.9 & 87.7 & 89.2 \\
RF & FastText & Binary & 90.3 & 88.5 & 89.4 \\
RF & FastText & None & 89.7 & 88.5 & 89.1 \\
RF & TF-IDF & Decimal & 90.8 & 89.0 & 89.8 \\
RF & TF-IDF & Binary & 91.3 & 88.3 & 89.7 \\
\textbf{RF} & \textbf{TF-IDF} & \textbf{None} & \textbf{90.4} & \textbf{91.4} & \textbf{90.9} \\
\hline

MLP & FastText & Decimal & 89.5 & 85.0 & 87.0 \\
MLP & FastText & Binary & 89.5 & 87.0 & 88.2 \\
MLP & FastText & None & 88.8 & 88.6 & 88.7 \\
MLP & TF-IDF & Decimal & 89.1 & 86.1 & 87.5 \\
MLP & TF-IDF & Binary & 89.1 & 84.4 & 86.4 \\
MLP & TF-IDF & None & 89.2 & 88.1 & 88.6 \\
\hline

Linear & FastText & Decimal & 83.2 & 82.5 & 82.9 \\
Linear & FastText & Binary & 87.9 & 81.9 & 84.4 \\
Linear & FastText & None & 87.1 & 85.5 & 86.3 \\
Linear & TF-IDF & Decimal & 84.1 & 65.5 & 69.1 \\
Linear & TF-IDF & Binary & 86.6 & 86.0 & 86.3 \\
Linear & TF-IDF & None & 88.0 & 88.2 & 88.1 \\
\hline
\end{tabular}
\end{center}
\end{table}

All configurations use the same split of the dataset with 60\% for the training set, 20\% for the validation set, and 20\% for the test set. Table~\ref{table:results} lists the results of all combinations of different variables. 
The first row lists the scores of the non ML baseline. 
The best ML configuration is the combination of \textit{RF}, \textit{TF-IDF}, and \textit{None} transformation, which obtained an F1-score of 90.9\%, with a strong gain of 23\% over the baseline.

\begin{figure}[htb]
  \centering
  \includegraphics[width=11.5cm]{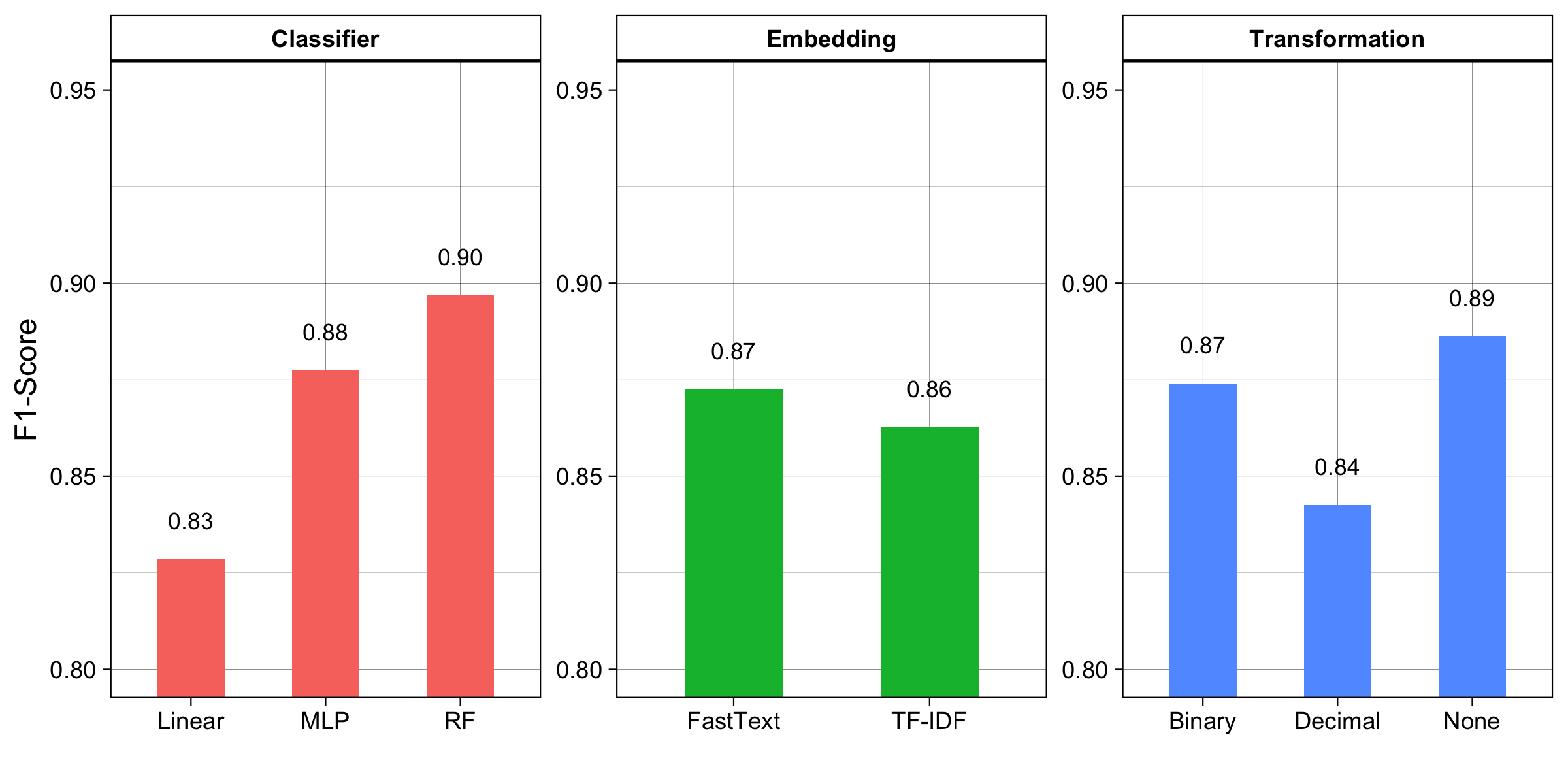}
  \caption{Average F1-score by aggregating different variables on test set}
  \label{fig3:f1_average}
\end{figure}

In Figure~\ref{fig3:f1_average}, to get a better understanding of the impact of each variable, we compute the average F1-score for each variable, aggregating on the other variables. 
On the one hand, we observe that the classifier \textit{RF} performs better than \textit{MLP} and \textit{Linear} while no transformation on feature values performs better than the other two strategies. 
On the other hand, we observe no significant difference between the two text embedding representations FastText and TF-IDF. 

In terms of errors, about 35\% of them arise from some required piece of information being not correctly extracted (\textit{partial, missing or incorrect}), for instance, a missed contract number will lead to an incomplete feature representation.
About 25\% of errors are due to confusion between amendments and other document types (such as appendixes) that exhibit naming patterns similar to amendments. 
20\% of errors result from the trained model being not able to capture the specific conventions followed by each organization, for instance, the contract referencing system. 
For the remaining 20\%, no reason could be clearly identified.

\section{Applications}
\label{sec:applicaiton}

Identifying the amendment relationship between a pair of document is a key step for real life CLM automation. For instance, we identified (and implemented) the following two concrete scenarios:

\begin{itemize}
    \item {\verb| Linked documents suggestion|}: When the user uploads a new document, the software can suggest a list of potentially linked documents to the user. 
    The user will need to simply verify and validate the suggested documents instead of searching or selecting manually the linked documents;
    \item {\verb| Automatic sorting|}: For new users, when they upload their contracts (generally a big volume) for the first time, this function could help them to structure their contract database by making explicit links between the master agreements and their amendments. 
\end{itemize}

In practice, some settings and thresholds on the prediction probability may differ according to the above-mentioned scenarios. 
In the first application, we wish to favor the \textit{recall} so as to suggest all possible linked documents: a relatively low threshold of probability will be sufficient. 
Additionally, a parameter \textit{top\_x} to limit the number of suggestions can be set, namely, to keep \textit{x} top best predictions.
Whereas in the second case, we prefer to set a higher threshold to guarantee a high \textit{precision}, in order to ensure that the automatically sorted documents are correct. 

\section{Conclusions and Future Works}
\label{sec:conclusion}
This paper addresses the problem of contract-amendment classification in CLM. 
We propose a distance and cross reference-based approach to build the features of a pair of documents and evaluate several configurations to classify the relationships. 
The best configuration outperforms 23\% in terms of F1-score compared to the baseline, which is a heuristic-based method without the application of machine learning techniques. 
The obtained results can be applied to different application scenarios in the CLM automation and bring real benefits to the final users. 

Based on the error analysis performed in Section \ref{sec:classification}, we plan working on the following aspects to improve the approach:

\begin{itemize}
    \item {\verb|Reinforce preprocessing|}: As 34\% of errors are related to the fact that the required information is not well extracted, the reinforcement of OCR and NER could improve these issues. 
    Furthermore, these improvements will contribute to other tasks in the whole CLM pipeline;
    
    \item {\verb|Train model by user|}: To capture the user preferences, training on the dataset of each user would help to make the model more customized and accurate;
    
    \item {\verb|Improve cross-reference detection|}: The current method checks the number of some shared keywords as a feature to detect cross document references. 
    However, we believe this can be improved with Named Entity Linking (\textit{NEL})~\cite{Hachey2013}, particularly a graph-based approach~\cite{hachey2011graph}.
\end{itemize}

\section*{Acknowledgment}
I thank Dr. \'Eric de la Clergerie, researcher at INRIA (team Alpage\footnote{\url{https://www.rocq.inria.fr/alpage-wiki/tiki-index.php?page=accueil}}) and the members of Data Science team at Hyperlex\footnote{\url{https://hyperlex.ai/}} for discussions and comments that improved this manuscript.

\bibliography{bibliography}

\begin{thebibliography}{11}
\expandafter\ifx\csname natexlab\endcsname\relax\def\natexlab#1{#1}\fi
\providecommand{\url}[1]{\texttt{#1}}
\providecommand{\href}[2]{#2}
\providecommand{\path}[1]{#1}
\providecommand{\DOIprefix}{doi:}
\providecommand{\ArXivprefix}{arXiv:}
\providecommand{\URLprefix}{URL: }
\providecommand{\Pubmedprefix}{pmid:}
\providecommand{\doi}[1]{\href{http://dx.doi.org/#1}{\path{#1}}}
\providecommand{\Pubmed}[1]{\href{pmid:#1}{\path{#1}}}
\providecommand{\bibinfo}[2]{#2}
\ifx\xfnm\relax \def\xfnm[#1]{\unskip,\space#1}\fi
\bibitem[{Nadeau and Sekine(2007)}]{Nadeau2007}
\bibinfo{author}{D.~Nadeau}, \bibinfo{author}{S.~Sekine},
\newblock \bibinfo{title}{A survey of named entity recognition and
  classification},
\newblock \bibinfo{journal}{Lingvisticae InvestigationesLingvisticæ
  InvestigationesLingvisticæ Investigationes. International Journal of
  Linguistics and Language Resources} \bibinfo{volume}{30}
  (\bibinfo{year}{2007}). \DOIprefix\doi{10.1075/li.30.1.03nad}.
\bibitem[{Yadav and Bethard(shed)}]{Yadav2019}
\bibinfo{author}{V.~Yadav}, \bibinfo{author}{S.~Bethard},
\newblock \bibinfo{title}{A survey on recent advances in named entity
  recognition from deep learning models},
\newblock \bibinfo{journal}{arXiv preprint arXiv:1910.11470}
  (\bibinfo{year}{2019, unpublished}).
\bibitem[{Pons and Latapy(2006)}]{Pons2006}
\bibinfo{author}{P.~Pons}, \bibinfo{author}{M.~Latapy},
\newblock \bibinfo{title}{Computing communities in large networks using random
  walks},
\newblock \bibinfo{journal}{Journal of Graph Algorithms and Applications}
  \bibinfo{volume}{10} (\bibinfo{year}{2006}).
  \DOIprefix\doi{10.7155/jgaa.00124}.
\bibitem[{Fortunato(2010)}]{Fortunato2010}
\bibinfo{author}{S.~Fortunato},
\newblock \bibinfo{title}{Community detection in graphs},
\newblock \bibinfo{journal}{Physics Reports} \bibinfo{volume}{486}
  (\bibinfo{year}{2010}). \DOIprefix\doi{10.1016/j.physrep.2009.11.002}.
\bibitem[{{Song} and {de la Clergerie}(2020)}]{Song2020}
\bibinfo{author}{F.~{Song}}, \bibinfo{author}{{\'E}.~{de la Clergerie}},
\newblock \bibinfo{title}{Clustering-based automatic construction of legal
  entity knowledge base from contracts},
\newblock in: \bibinfo{booktitle}{2020 IEEE International Conference on Big
  Data (Big Data)}, \bibinfo{year}{2020}, pp. \bibinfo{pages}{2149--2152}.
  \DOIprefix\doi{10.1109/BigData50022.2020.9378166}.
\bibitem[{Sch{\"u}tze et~al.(2008)Sch{\"u}tze, Manning, and
  Raghavan}]{schutze2008introduction}
\bibinfo{author}{H.~Sch{\"u}tze}, \bibinfo{author}{C.~D. Manning},
  \bibinfo{author}{P.~Raghavan}, \bibinfo{title}{Introduction to information
  retrieval}, volume~\bibinfo{volume}{39}, \bibinfo{publisher}{Cambridge
  University Press Cambridge}, \bibinfo{year}{2008}.
\bibitem[{Powers(2020)}]{powers2020evaluation}
\bibinfo{author}{D.~M. Powers},
\newblock \bibinfo{title}{Evaluation: from precision, recall and f-measure to
  roc, informedness, markedness and correlation},
\newblock \bibinfo{journal}{arXiv preprint arXiv:2010.16061}
  (\bibinfo{year}{2020}).
\bibitem[{Rajaraman and Ullman(2011)}]{rajaraman_ullman_2011}
\bibinfo{author}{A.~Rajaraman}, \bibinfo{author}{J.~D. Ullman},
  \bibinfo{title}{Data Mining}, \bibinfo{publisher}{Cambridge University
  Press}, \bibinfo{year}{2011}, p. \bibinfo{pages}{1–17}.
  \DOIprefix\doi{10.1017/CBO9781139058452.002}.
\bibitem[{Bojanowski et~al.(2017)Bojanowski, Grave, Joulin, and
  Mikolov}]{bojanowski2017enriching}
\bibinfo{author}{P.~Bojanowski}, \bibinfo{author}{E.~Grave},
  \bibinfo{author}{A.~Joulin}, \bibinfo{author}{T.~Mikolov},
\newblock \bibinfo{title}{Enriching word vectors with subword information},
\newblock \bibinfo{journal}{Transactions of the Association for Computational
  Linguistics} \bibinfo{volume}{5} (\bibinfo{year}{2017})
  \bibinfo{pages}{135--146}.
\bibitem[{Hachey et~al.(2013)Hachey, Radford, Nothman, Honnibal, and
  Curran}]{Hachey2013}
\bibinfo{author}{B.~Hachey}, \bibinfo{author}{W.~Radford},
  \bibinfo{author}{J.~Nothman}, \bibinfo{author}{M.~Honnibal},
  \bibinfo{author}{J.~R. Curran},
\newblock \bibinfo{title}{Evaluating entity linking with wikipedia},
\newblock volume \bibinfo{volume}{194}, \bibinfo{year}{2013}.
  \DOIprefix\doi{10.1016/j.artint.2012.04.005}.
\bibitem[{Hachey et~al.(2011)Hachey, Radford, and Curran}]{hachey2011graph}
\bibinfo{author}{B.~Hachey}, \bibinfo{author}{W.~Radford},
  \bibinfo{author}{J.~R. Curran},
\newblock \bibinfo{title}{Graph-based named entity linking with wikipedia},
\newblock in: \bibinfo{booktitle}{International conference on web information
  systems engineering}, \bibinfo{organization}{Springer}, \bibinfo{year}{2011},
  pp. \bibinfo{pages}{213--226}.

\end{thebibliography}
\end{document}